\documentclass[runningheads]{llncs}

\usepackage{eccv}

\usepackage{eccvabbrv}
\usepackage{array}

\usepackage{graphicx}
\usepackage{booktabs}
\usepackage{amsmath}
\usepackage{mathtools, bbm}
\usepackage{amssymb}
\usepackage{amsfonts} 
\usepackage{nicefrac}       % compact symbols for 1/2, etc.
\usepackage{multirow}
\usepackage{siunitx}
\usepackage{enumerate}
\usepackage{wrapfig}
\usepackage{pifont}
\usepackage{caption,stackengine}

\usepackage[accsupp]{axessibility}  % Improves PDF readability for those with disabilities.
\usepackage{hyperref}

\usepackage{orcidlink}

\AtBeginDocument{\DeclareSIUnit{\kWh}{kWh}}

\newcommand{\xmark}{\ding{55}}%
\newcommand{\minisection}[1]{\vspace{0.03in} \noindent {\bf #1}}

\begin{document}

%\title{Green AI in Incremental Learning: An Empirical Overview} 
% \title{Is Continual Learning always Green ? An Empirical Evaluation with Vision Foundation Models}
\title{How green is continual learning, really? Analyzing the energy consumption in continual training of vision foundation models}
%\title{How green is continual learning, really? \\ An empirical analysis on vision foundation models}
\titlerunning{How green is continual learning, really?}

%\author{Tomaso Trinci\inst{1}\orcidlink{0000-1111-2222-3333} \and
%Simone Magistri\inst{2,3}\orcidlink{1111-2222-3333-4444} \and
%Roberto Verdecchia\inst{3}\orcidlink{2222--3333-4444-5555} \and Andrew D. Bagdanov\inst{3}\orcidlink{2222--3333-4444-5555}}

\author{Tomaso Trinci\orcidlink{0000-0002-4052-1930} \and
Simone Magistri\orcidlink{0000-0002-0520-8463}  \and
Roberto Verdecchia\orcidlink{0000-0001-9206-6637} \and 
Andrew D. Bagdanov\orcidlink{0000-0001-6408-7043}}

%Andrew D. Bagdanov\inst{1}\orcidlink{0000-0001-6408-7043}
%simo: \inst{1}\orcidlink{0000-0002-0520-8463}
% Roberto: https://orcid.org/0000-0001-9206-6637

\authorrunning{T.~Trinci et al.}

\institute{Department of Information Engineering, University of Florence, Italy \\
\email{name.surname@unifi.it}}

%\institute{Princeton University, Princeton NJ 08544, USA \and
%Springer Heidelberg, Tiergartenstr.~17, 69121 Heidelberg, Germany

%

\maketitle

%%%%%%%%%%%%%%%%%%%%%%%%%%%%%%%%%%%%%%%%%%%%%%%%%%%%%%%%%%%%%%%%%%%% ABSTRACT

\begin{abstract}
With the ever-growing adoption of AI, its impact on the environment is no longer negligible. Despite the potential that continual learning could have towards Green AI, its environmental sustainability remains relatively uncharted. In this work we aim to gain a systematic understanding of the energy efficiency of continual learning algorithms. To that end, we conducted an extensive set of empirical experiments comparing the energy consumption of recent representation-, prompt-, and exemplar-based continual learning algorithms and two standard baseline (fine tuning and joint training) when used to continually adapt a pre-trained ViT-B/16 foundation model. We performed our experiments on three standard datasets: CIFAR-100, ImageNet-R, and DomainNet. Additionally, we propose a novel metric, the \textit{Energy NetScore}, which we use measure the algorithm efficiency in terms of energy-accuracy trade-off. Through numerous evaluations varying the number and size of the incremental learning steps, our experiments demonstrate that different types of continual learning algorithms have very different impacts on energy consumption during both training and inference. Although often overlooked in the continual learning literature, we found that the energy consumed during the inference phase is crucial for evaluating the environmental sustainability of continual learning models.
\end{abstract}

%%%%%%%%%%%%%%%%%%%%%%%%%%%%%%%%%%%%%%%%%%%%%%%%%%%%%%%%%%%%%%%%%%%% INTRO

\section{Introduction}
\label{sec:introduction}

The widespread adoption of Artificial Intelligence for real-world applications has been driven by the ability of deep learning to solve increasingly complex problems in various fields, such as computer vision and natural language processing~\cite{LeCun2015}. The growing demand for higher-performing models has led to the development of large models pre-trained on massive datasets, such as Llama~\cite{touvron2023llama} and CLIP~\cite{dosovitskiy2021an}, whose training requires intensive hardware resources. Such models, based on Language and Vision Transformer architectures~\cite{NIPS2017_3f5ee243, dosovitskiy2021an} and commonly referred to as \textit{foundation models}~\cite{bommasani2021opportunities}, can reach hundreds of billions of parameters and perform trillions of operations. This scale has prompted \textit{Green AI} researchers to address the environmental concerns associated with them~\cite{JMLR:v24:23-0069}.  

Current best practices involve adapting foundation models through transfer learning, or \textit{fine-tuning}, to solve specific tasks. However, the standard learning paradigm of these approaches is largely \textit{static}. When models must be updated to learn new tasks or to improve their performance with additional data, the preferred solution is \textit{joint-training} on both new and old data. This may offer optimal performance, but demands increasing computational resources as the number of tasks or the volume of data grows. Conversely, sequentially fine-tuning on novel tasks requires fewer training resources but can lead to a rapid decrease in performance on previous tasks, a phenomenon known as \textit{catastrophic forgetting}~\cite{9349197, MCCLOSKEY1989109}
 
Continual Learning (CL) aims at enabling deep learning models to continuously learn from new data while mitigating catastrophic forgetting~\cite{verwimp2024continual}. This enables model update even when old data disappear due to \textit{privacy concerns} and requires \textit{fewer training computational resources}, reducing environmental impact and resource costs~\cite{8107520, pmlr-v78-lomonaco17a}. These benefits make CL appealing for applications where privacy and efficiency are major concerns~\cite{Shaheen2022, 10157835, MAGISTRI202482, HURTADO2023200251}. The development of efficient training strategies has been the primary motivation for CL approaches aimed at mitigating forgetting in large, hardware-intensive pre-trained foundation models~\cite{wang2022learning, zhou2024expandable, mcdonnell2024ranpac}. Surprisingly, despite the efficiency promises of CL, and the non-negligible carbon footprint of foundation model training, a systematic empirical evaluation of the sustainability of CL appears to be missing in the literature. Specifically, there is a lack of analysis on the additional training complexities introduced to address catastrophic forgetting and how current CL methods compare to joint training in terms of energy usage. Additionally, the complexity these CL methods add to the models inference phase has been largely overlooked in the CL research community.

\begin{figure}[t]
    \centering
    \includegraphics[width=.90\textwidth]{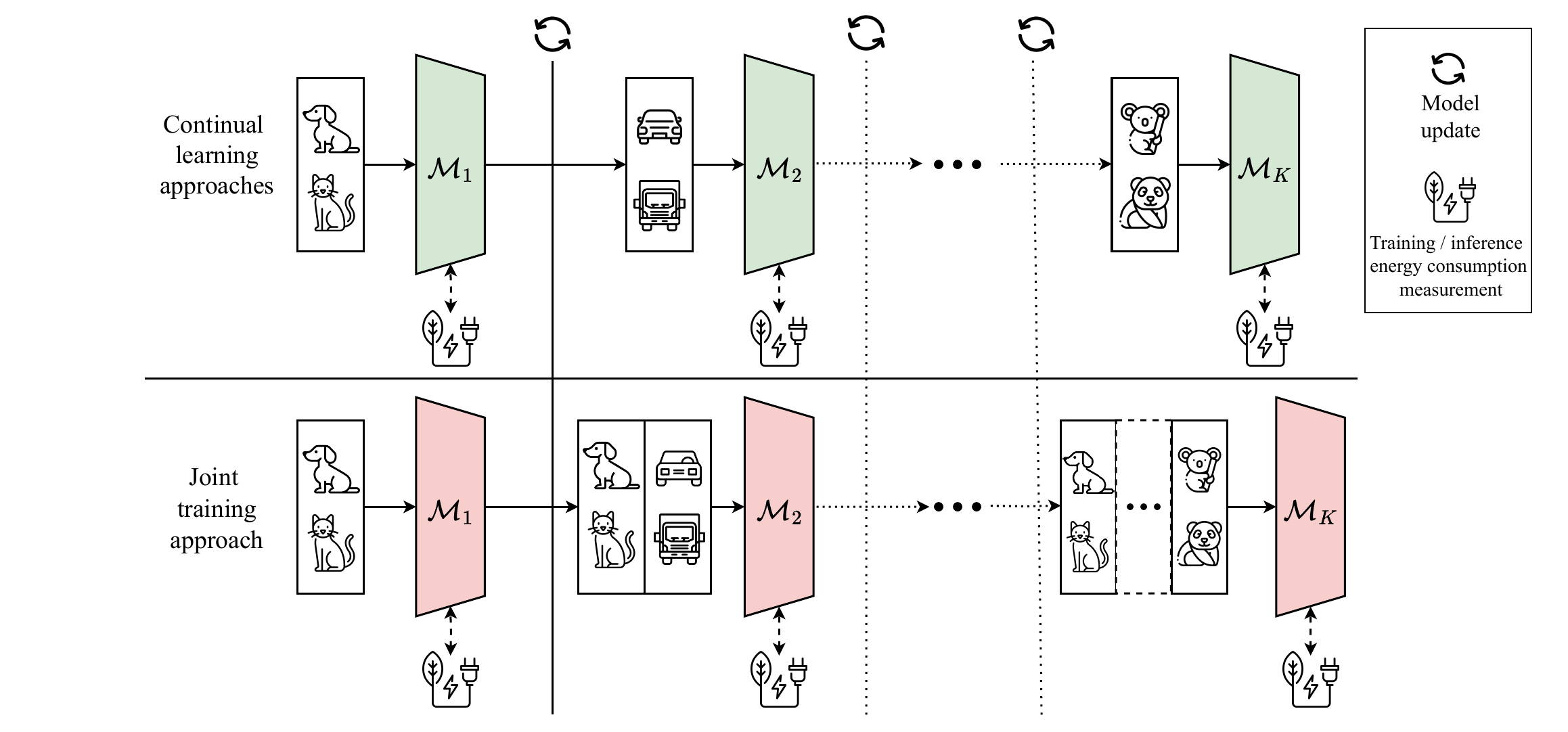}
    \caption{In continual learning a model $\mathcal{M}_k$ (in green) learns and adapts using only current data without forgetting previous information. Conversely, the joint incremental training strategy (in red) uses both previous and current data, leading to comprehensive learning but higher computational and storage costs. In this work we aim to understand the impact on the energy consumption of model $\mathcal{M}_k$ when trained following different CL approaches and how they compare to joint incremental training.}
    \label{fig:research_process}
\end{figure}

In this paper we present an extensive empirical investigation in the energy consumption of CL with pre-trained models, focusing on vision foundation models due to its prominence in CL literature. Our study aims to determine whether CL methods are energy efficient during training and inference phases, and hence whether (and to what degree) they render CL sustainable in real-world applications. Specifically, we measure the energy costs associated with CL techniques applied to pre-trained vision models and compare them to the energy required to completely retrain a model for each new task (see Figure~\ref{fig:research_process}). Moreover, we propose a new metric called the \textit{Energy NetScore} that measures overall algorithm efficiency in terms of energy-accuracy trade-off. By exploring the intersection of vision foundation models, continual learning, and Green AI, we aim to address a gap in current research and to evaluate how effective CL is at saving energy.

This research empirically answers key research questions regarding the energy consumption of CL applied to pre-trained vision foundation models:
\begin{itemize}
    \item[\textbullet] \textit{How do CL approaches compare in training energy consumption?}
    \item[\textbullet] \textit{How do training energy costs scale with data and model updates?}
    \item[\textbullet] \textit{How do CL approaches compare in Energy NetScore?}
    \item[\textbullet] \textit{How does the CL strategy impact inference energy consumption?}
\end{itemize}

%%%%%%%%%%%%%%%%%%%%%%%%%%%%%%%%%%%%%%%%%%%%%%%%%%%%%%%%%%%%%%%%%%%% RELATED WORK

\section{Related Work}
\label{sec:related_work}
In recent years research into the environmental impact of AI systems has gained increasing momentum~\cite{verdecchia2023systematic}. Among the many works on this topic, a considerable portion of Green AI studies propose solutions to optimize the environmental impact of AI through different strategies. Green AI solutions often consider hyperparameter tuning~\cite{de2021hyperparameter}, deployment~\cite{tao2020challenges}, data-centric~\cite{verdecchia2022data}, and trade-off between precision and energy consumption strategies~\cite{zhang2018exploring}. 

A smaller set of Green AI works instead present observational studies, i.e. inquiries aimed to solely assess, rather than optimize, the environmental sustainability of AI. Examples of observational studies include the comparison of different deep learning frameworks in terms of energy efficiency~\cite{georgiou2022green}, understanding the impact of hyperparameter tuning on power consumption~\cite{de2021hyperparameter}, and monitoring carbon intensity of AI algorithms in cloud environments~\cite{dodge2022measuring}. Our study falls in the Green AI observational study category, as it presents an empirical evaluation of the energy efficiency of continual training of vision foundation models.

In terms of architectures considered in Green AI studies, we note that neural networks are by far most investigated~\cite{verdecchia2023systematic}. Studies falling under this category cover a vast and heterogeneous spectrum of topics, ranging from the impact of precision quantization on energy efficiency~\cite{hashemi2017understanding}, power capping techniques~\cite{krzywaniak2022gpu} and deployment strategies in multi-GPU environments~\cite{castro2019energy} to energy consumption prediction approaches~\cite{rodrigues2018synergy}. In such a variegated set of arguments connected to the environmental sustainability of neural networks, our study positions itself by exploring a niche that has to date remained largely unexplored, namely Green AI in Continual Learning, as further discussed below.

To the best of our knowledge, only a single study takes into account the environmental sustainability of CL. In their work, Chavan et al.~\cite{chavan2023towards} present an observational study assessing the energy efficiency of four CL algorithms in an industrial setting. They argue that in such scenarios it preferable to alternate CL techniques with complete retraining when certain thresholds of acceptability are reached. In contrast, we are interested in comparing different continual learning methods and understanding how various solutions to mitigate forgetting impact energy consumption, both during training and inference phases when adapting vision foundation models to a sequence of incremental tasks.

We believe that our study presents the most extensive evaluation to date of continual learning algorithms in terms of environmental sustainability. In the next section we describe the specific scenario we consider and the CL approaches of vision foundation models we consider in our evaluation.

%%%%%%%%%%%%%%%%%%%%%%%%%%%%%%%%%%%%%%%%%%%%%%%%%%%%%%%%%%%%%%%%%%%% CL BACKGROUND

\section{Continual Learning of Vision Foundation Models}
\label{sec:cl_background}
In this section we introduce the key concepts of CL applied to vision foundation models, which are at the basis of our empirical evaluation. 

\subsection{Preliminaries}
CL addresses the challenge of training a model on a sequence of non-stationary data without the need to retrain on all previously used data, a process known as joint incremental training, while also preventing forgetting. Traditionally, CL has been evaluated using Convolutional Neural Network (CNN) models~\cite{9891836, 10.1007/978-3-030-58565-5_6, Zhu_2021_CVPR, Goswami_2024_CVPR, magistri2024elastic}. However, the impressive performance of large vision foundation models pre-trained on extensive datasets, such as the Vision Transformer (ViT)~\cite{dosovitskiy2021an}, has led recent incremental learning studies to shift their focus toward CL with these pre-trained models as opposed to training from scratch~\cite{wang2022learning, smith2023coda, goswami2024fecam, COSSU2024106492}.

%ViT models are vision models based on the widely-used Transformer architecture~\cite{NIPS2017_3f5ee243}. In ViT, the input image is partitioned into patches, which are then processed as a sequence of tokens. Additionally, a learnable class token is concatenated to this sequence of patches, forming the complete input sequence. Specifically, a stack of layers, alternating multi-head attention and multi-layer perceptron operations, processes the input sequence, and generates an embedding sequence. The final embedding of the input image $x$ is represented by the embedding of the class token after the last layer, denoted as \(z_{\text{cls}}(x; \Theta)\), where \(\Theta\) are the parameters of the model. This final embedding is then used by a classifier to produce predictions.

ViT models are vision models based on the Transformer architecture~\cite{NIPS2017_3f5ee243}. In ViT, an image is divided into patches, treated as a sequence of tokens, and a learnable class token is added. The input sequence is processed by alternating layers of multi-head attention and multi-layer perceptrons. The final embedding of the input image $x$, represented by the class token after the last layer as $z_{\text{cls}}(x; \Theta)$ with $\Theta$ model's parameters, is used by a classifier to generate predictions.

In the context of CL, a pre-trained ViT model $\mathcal{M}_t$ is sequentially trained on $K$ tasks, typically image classification tasks, each characterized by a disjoint set of classes. We can define the sequence of task used to train the model as \(\mathcal{D}_K = \{\mathcal{X}_t, \mathcal{Y}_t\}_{t=1}^K\), where \(\mathcal{X}_t\) and \(\mathcal{Y}_t\) are the sets of images and classification labels for task \(t\), respectively. The output of the model after task $t$ for an input $x$ is the composition of its class token $z_{\text{cls}}(x; \Theta_t)$, depending on parameters $\Theta_t$, and a classifier with parameters $W_t$, which are sequentially updated on new tasks. The simplest classifier is a single linear layer (or classification head) followed by a softmax activation function, resulting in the model output:
\begin{equation}
    \mathcal{M}_t(x; \Theta_t, W_t) \equiv p(y \mid x; \Theta_t, W_t) = \text{softmax}(W_t^{\top} z_{\text{cls}}(x; \Theta_t)).
\end{equation}
Training $\mathcal{M}_t$ only on the current task data $t$ (i.e., fine-tuning) is more efficient than re-training on all previous task data. However, this approach can lead to what is known as \textit{catastrophic forgetting}: when a model forgets the initial task after learning one or more new tasks~\cite{MCCLOSKEY1989109}. The causes of catastrophic forgetting can be found in: (i) \textit{weight/activation drift}, where training on a new tasks modifies crucial weights for previous tasks; (ii) \textit{task-recency bias}, where the linear classifier output becomes biased towards new task classes, leading to uncalibrated predictions for previous classes; and (iii) \textit{inter-task confusion}, where the final classifier, trained only on the last task's classes, has suboptimal decision boundaries for the previous tasks' classes~\cite{9915459}. 

CL can be viewed as a collection of training techniques aimed at mitigating catastrophic forgetting. In the following section, we give insights into specific methodologies that we utilized in our analysis.

\subsection{Continual Learning Approaches}
\label{sec:cl_appr}
Continual learning approaches can be broadly categorized into Class-Incremental Learning (class-IL), where the model must distinguish among an increasing number of classes; Task-Incremental Learning (task-IL), where the model learns distinct tasks in sequence; and Domain-Incremental Learning (domain-IL), where the model is trained on a fixed task but with varying input domains~\cite{van2022three}.

Continual learning methods can be further distinguished as either \textit{exemplar-based} or \textit{exemplar-free}. The former stores and replays a subset $\mathcal{E}_{t-1}$ of previous task data, known as \textit{exemplars}, when learning new tasks~\cite{rebuffi2017icarl}, while the latter relies only on the current task data~\cite{wang2022learning}.

In this paper, we focus on the class-IL setting, the most explored in CL with pre-trained vision models, highlighting the key features and energy impacts of both exemplar-free and exemplar-based methods.

\minisection{Exemplar-based. } iCaRL~\cite{rebuffi2017icarl} is an exemplar-based method, originally designed for incremental learning with CNNs, but recently adapted for pre-trained ViT architecture~\cite{wang2022learning, zhou2024expandable}. % iCaRL fine-tunes the entire ViT with both exemplars and current task data using the following loss function:%a cross entropy loss and knowledge distillation~\cite{hinton2015distilling}: 
%\begin{equation}
%    \mathcal{L}_t^\text{iCaRL} =  \mathcal{L}^{\text{ce}}_t(\mathcal{X}_t, \mathcal{E}_{t-1})+ \mathcal{L}^{\text{KD}}_t (\mathcal{X}_t, \mathcal{E}_{t-1}). 
%\end{equation}
iCaRL fine-tunes the entire ViT with both exemplars and current task data making use of a cross entropy loss and knowledge distillation as regularizer~\cite{hinton2015distilling}.
%where $\mathcal{L}^{\text{ce}}_t$ and $\mathcal{L}^{\text{KD}}_t$ are respectively the cross entropy and the knowledge distillation losses evaluated on current task data and exemplars.
In particular, the regularizer aligns the output of the current model $\mathcal{M}_t$ with those of the previous model $\mathcal{M}_{t-1}$ to mitigate activation drift. At the end of each task, iCaRL mitigates task-recency bias by computing the mean of class tokens and using nearest mean classification for predictions, eliminating the need for the linear classifier $W_t$.

MEMO~\cite{zhou2022model} is an exemplar-based method, which dynamically expand the network as the number of tasks increases. In MEMO, the initial layers of the ViT are frozen, while the final layers are duplicated and trained separately for each task. Like iCaRL, MEMO uses exemplars to mitigate the activation drift.

\minisection{Exemplar-free.} Recent exemplar-free approaches for pre-trained models include the \textit{prompt-based} methods. These methods keep the pre-trained feature extractor frozen and inject a few learnable parameters, called \textit{prompts}, before the multi-head self-attention layers. These prompts guide the pre-trained model in learning new tasks by adapting the feature representation. The output of an incremental ViT model $\mathcal{M}_t$ using prompts is:
 \begin{equation}
    \mathcal{M}_t(x; \Theta_0, W_t, P^1_t,\ldots,P^M_t) \equiv \text{softmax} (W_t^{\top} z^L_{\text{cls}}(x; \Theta_0, P^1_t,\dots,P^M_t)),
\end{equation}
where $\Theta_0$ are the pre-trained frozen weights and $P^1_t,\dots,P^M_t$ are the learnable prompts updated across tasks. At training and test time, these prompts are selected from a pool $\mathcal{P}$ %, with $\vert\mathcal{P}\vert > M$, 
using a query function calculated on the input $x$. A common approach is to use the class token $z^L_{\text{cls}}(x; \Theta_0)$  as query function~\cite{wang2022learning, smith2023coda}. %The cosine similarity between this token and each pool element is computed and the top $M$ nearest prompts in the pool which are selected and used with the input $x$ to get the model output $\mathcal{M}_t$.

Prompt-based algorithms vary in the prompt selection processes, in the number of prompts $M$ used, and in the injection points into the model. Learning to Prompt (L2P)~\cite{wang2022learning} appends learnable prompts to the embedding input sequence. DualPrompt~\cite{wang2022dualprompt} injects \textit{general} prompts, shared across tasks, and \textit{expert} prompts, which are task-specific, in different points along the backbone. CODA-Prompt~\cite{smith2023coda} modifies DualPrompt by splitting the prompts into \textit{prompt components} and it uses a weighted summation with learnable weights to determine which components to utilize during inference.

Another category of exemplar-free methods is the \textit{representation-based} methods, which aim to maximize the model's ability to generate high-quality features from the pre-trained model, even for data not specifically trained on~\cite{sun2023pilot}.

EASE~\cite{zhou2024expandable} freezes the pre-trained feature extractor and adapts it using task-specific learnable adapters, then uses a prototype-based classifier for predictions and estimates representation drift of old prototypes as new tasks are added. At inference time, images are passed through multiple adapters. SimpleCIL~\cite{zhou2023revisiting}, similarly to~\cite{janson2022a}, computes the class means of the feature representations without any training, and uses these representations as a weight matrix for a final linear classifier. RanPAC~\cite{mcdonnell2024ranpac}, like some recent CL approaches~\cite{panos2023first,goswami2024fecam}, employs a \textit{first session adaptation} of the backbone, where the ViT is adapted during the first task by updating a few learnable weights through parameter-efficient fine-tuning. Additionally, it enhances feature space via random projection to improve class separability. The final classification is based on a metric derived from the Gram Matrix of features and class means. For subsequent tasks, the ViT is frozen, and only the Gram Matrix and class means are updated to account for new classes.

\minisection{Efficiency Considerations}. The previous sections hint at the strengths and weaknesses of the various approaches in terms of efficiency, which we aim to quantify. Exemplar-free methods are more training-efficient than exemplar-based methods (i.e., iCaRL and MEMO) as they avoid sample replay during the current task training~\cite{magistri2024elastic}. However, these methods come with their own complexities. For example, prompt-based methods require double inferences: one for prompt selection and another for the forward pass. EASE increases architecture size and necessitates multiple forward passes at inference time. RanPAC significantly expands the feature space dimensionality, resulting in a large final classifier. In the next section, we document how we empirically analyze the energy consumption of the different CL approaches, which is often taken for granted by CL literature~\cite{wang2022learning,zhou2024expandable}. Additionally, we introduce a novel metric to evaluate the trade-off between performance and efficiency in these approaches.

%%%%%%%%%%%%%%%%%%%%%%%%%%%%%%%%%%%%%%%%%%%%%%%%%%%%%%%%%%%%%%%%%%%% METHODOLOGY

\section{Methodology}
\label{sec:methodology}
\begin{figure}[t]
    \centering
    \includegraphics[width=0.9\textwidth]{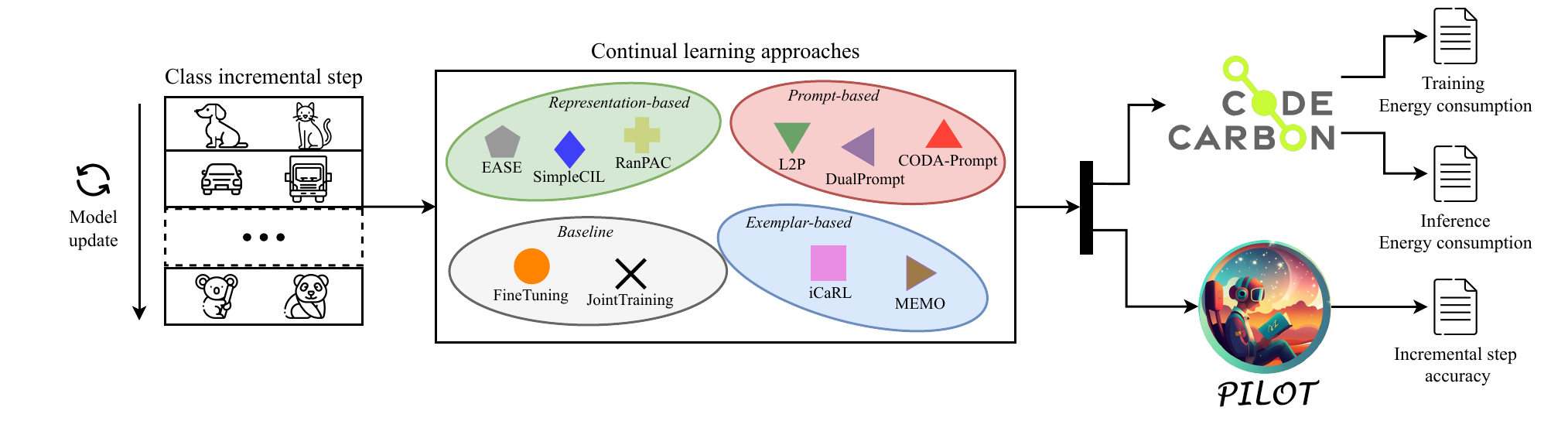}
    \caption{Overview of our experimental methodology. PILOT~\cite{sun2023pilot} is the framework that implements the CL approaches measuring the accuracy over incremental training steps, while CodeCarbon~\cite{benoit_courty_2024_11171501} evaluates energy consumption during training and inference.}
\label{fig:methodology}
\end{figure}

The previous section outlines the inherent complexity entailed by the multifaceted and heterogeneous characteristics of CL approaches. To tackle the challenges of experimenting with CL, our research methodology is based on directly measuring energy consumption of approaches, both during training and inference, followed by a thorough analysis of accuracy and efficiency trade-offs. Specifically, we aim to quantitatively understand:
\begin{itemize}
    \item[\textbullet] The impact of replaying old data in terms of energy cost relative to performance gains in exemplar-based approaches, identifying the incremental scenarios (if they exist) in which is most beneficial.
    \item[\textbullet] The energy savings achievable by utilizing a small number of trainable parameters in prompt-based methods, balanced against the additional cost of the double inference required for prompt selection.
    \item[\textbullet] The benefits of heavily relying on a pre-trained backbone without adaptation, or with a first task adaptation, weighted against the challenges posed by significant domain shifts or potential increases in inference costs.
\end{itemize}
To achieve our research goal, we use two Python libraries: the CL framework PILOT~\cite{sun2023pilot} and CodeCarbon~\cite{benoit_courty_2024_11171501}, a package for measuring energy consumption. Figure~\ref{fig:methodology} provides a high-level overview of our experimental methodology, detailing both the methods and the empirical output data collected.

In the following sections we explain in detail the procedure we leverage for measuring the energy consumption, supported by the introduction of a novel metric to evaluate models by considering both its performance and energy usage.

\subsection{Measuring Energy Consumption}
To monitor the energy footprint of the CL approaches, we used CodeCarbon~\cite{benoit_courty_2024_11171501}, a Python package that estimates the energy usage of the key hardware components, namely memory, CPU, and GPU. Memory consumption ($E_{\text{RAM}}$) is estimated using a model that depends on the amount of allocated memory, while the CPU energy usage ($E_{\text{CPU}}$), CodeCarbon utilizes \textit{Intel's RAPL} (Running Average Power Limit) interface. %RAPL allows software to monitor and control the energy usage of the processor and its components.
For GPU ($E_{\text{GPU}}$), it uses the \textit{pynvml} library, which is specific to NVIDIA GPUs. The energy consumption reported by RAPL and pynvml reflects the consumption of the respective devices and not just the observed process, so experiments must be conducted in a controlled environment~\cite{heguerte2023estimate}.

Throughout this paper, the reported energy values will always refer to the \textit{total energy consumed} during an experiment in kWh, i.e., $E = E_{\text{RAM}} + E_{\text{CPU}} + E_{\text{GPU}}$. This is not a limitation of our analysis, as we found that the distribution of energy consumption among the components remains consistent regardless of the dataset and method analyzed. Specifically, GPUs are responsible for approximately 79\% of the total energy consumption during training, while CPUs and RAM account for 15\% and 6\%, respectively. These numbers are in line with those reported in~\cite{hodak2019towards}. In practical terms, for each approach, we systematically isolate the functions responsible for the training loop and evaluation and collect the respective consumption data for each incremental step.

\subsection{The Energy NetScore}
Many different metrics can evaluate neural networks based on accuracy and architectural and computational complexity. Canziani et al. propose the \textit{information density} representing the accuracy-to-parameter ratio of a network~\cite{canziani2016analysis}. Building on this metric, Wong introduces the \textit{NetScore}, which keep in consideration the number of multiply–accumulate (MAC) operations needed for inference into this ratio~\cite{wong2019netscore}. The NetScore was further elaborated and adapted for CL to include other measurment such as training time, memory occupancy and the number of backward operations~\cite{Hayes_2022_Embedded_CL,harun2023efficient}. 

%In this paper, we propose to represent the architectural and computational complexity by directly considering the total energy consumed by the model during the experiment. This consumption, combined with accuracy, reflects the trade-off between performance and efficiency and will be used to rank the analyzed approaches. Therefore, following the approach similar to~\cite{wong2019netscore}, we define the \textit{Energy NetScore} of the model $\mathcal{M}$ as 

In this paper we propose representing architectural and computational complexity by considering the energy consumed by the model during an entire experiment. This, combined with accuracy, defines what we call the \textit{Energy NetScore} of the model $\mathcal{M}$, denoted as $\Omega(\mathcal{M})$. It measures the trade-off between performance and efficiency and will be used to rank the analyzed approaches. Specifically, following a similar approach to~\cite{wong2019netscore}, we define $\Omega(\mathcal{M})$ as 
\begin{equation}
\label{eqn:our_netscore}
    \Omega(\mathcal{M}) \vcentcolon = 20 \log \left( \dfrac{A(\mathcal{M})^{\alpha}}{E(\mathcal{M})^{\beta}} \right),
\end{equation}
where $\alpha$ and $\beta$ controlling the influence of accuracy $A(\mathcal{M})$ and energy consumed $E(\mathcal{M})$ of the model on the score, respectively. Although in this paper we applied this evaluation in the context of CL, we emphasize that the definition of the Energy NetScore is quite general and can be used to evaluate the trade-off between performance and consumption in any scenario.

%%%%%%%%%%%%%%%%%%%%%%%%%%%%%%%%%%%%%%%%%%%%%%%%%%%%%%%%%%%%%%%%%%%% EXPERIMENTAL SETUP

\section{Experimental Setup}
Here we document the benchmarks, metrics, and experimental settings we used (see Appendix~\ref{app:details} for further details).

\minisection{Benchmarks. }% Since all the tested approaches used a pretrained backbone initially trained on ImageNet-21K~\cite{ridnik2021imagenetk} and subsequently fine-tuned on ImageNet-1K~\cite{deng2009imagenet} , evaluating methods on these datasets or their subsets is not meaningful. 
Following~\cite{wang2022learning, zhou2024expandable, mcdonnell2024ranpac}, we evaluated the methods on CIFAR-100~\cite{CIFAR100}, ImageNet-R~\cite{hendrycks2021many}, and DN4IL~\cite{gowda2023dual}, a balanced subset of the DomainNet dataset. For CIFAR-100, which consists of 100 classes and 60,000 images, we used a 10-step scenario where the classes are equally split into 10 tasks. For ImageNet-R, which has 200 classes and 30,000 images, we considered both 10-step and 20-step scenarios with 3,000 and 1,500 examples per task, respectively. DN4IL consists of six domains, each with 100 classes, and a total of 85,000 images. For this benchmark, each domain (i.e. quickdraw, infograph, etc) is treated as a different task, resulting in a 6-step scenario and it is used to evaluate the performance when large domain shifts occur.

\minisection{Metrics.}
CL approaches are evaluated by measuring their accuracy after each task. We refer to this quantity as \textit{per-step accuracy} ($A_k$). We denote $E_k$ the total energy consumption upon task $k$, measured in kWh. 
% It is the sum of the consumption of each device, as specified in Eq.~\eqref{eqn:energy}. 
For the Energy NetScore $\Omega_k$ at task $k$, defined in Eq.~\eqref{eqn:our_netscore}, following \cite{harun2023efficient}, we set $\alpha = 2$ and $\beta = 0.125$. 

\minisection{Training Details.}
All approaches start from a ViT-B/16~\cite{dosovitskiy2021an} pretrained on ImageNet-21K and fine-tuned on ImageNet-1K as the backbone. For hyperparameters we followed those suggested in PILOT. To ensure a fair comparison between approaches, we maintained a fixed batch size of 64 and set the number of epochs to 20 for each task. JointTraining is not implemented in PILOT and we follow the settings suggested in~\cite{mcdonnell2024ranpac}. For ICaRL and MEMO, we use 20 exemplars per class as memory scenarios, as is common in the literature~\cite{Kang2022afc}.

\minisection{Hardware. }
The experiments were conducted on a Linux server with an Intel\textregistered~Core\textsuperscript{TM} i5-10600KF CPU, an NVIDIA RTX\textsuperscript{TM} 3060 GPU, and 64GB of DDR3 RAM. We ensure CodeCarbon recognizes both CPU and GPU, enabling real-time energy consumption data collection without approximations.

%%%%%%%%%%%%%%%%%%%%%%%%%%%%%%%%%%%%%%%%%%%%%%%%%%%%%%%%%%%%%%%%%%%% EXPERIMENTAL RESULT

\section{Experimental Results}
\label{sec:Results}
In this section, we present and discuss the results of our empirical experiment, with a primary focus on addressing our research questions (see Section~\ref{sec:introduction}).

\begin{figure}[t]
    \centering
    \includegraphics[width=0.9\textwidth]{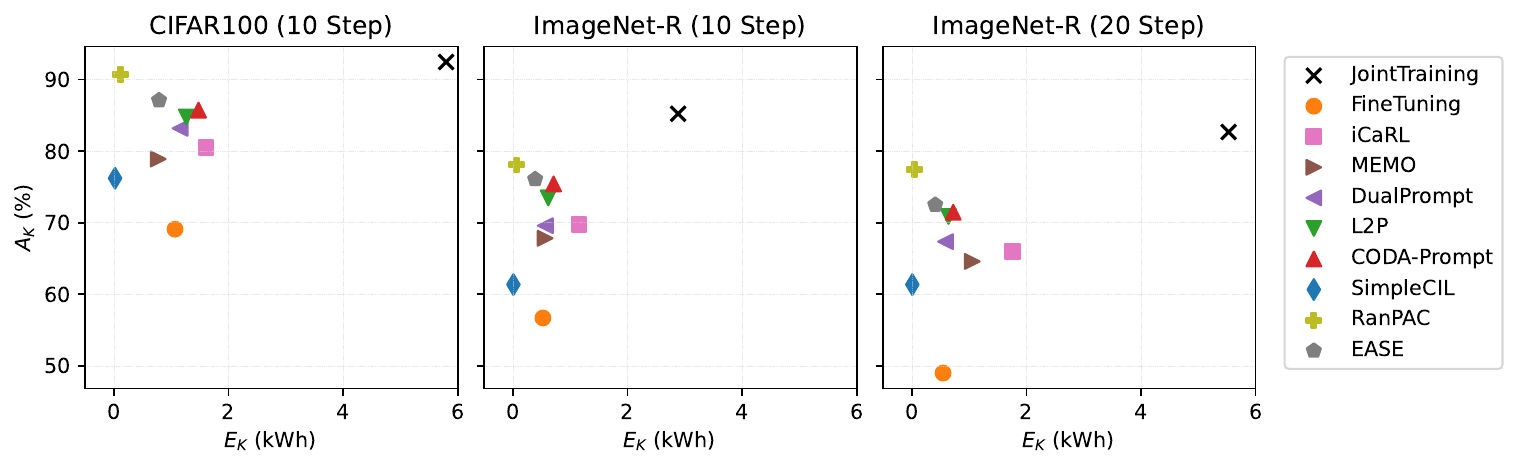}
    \caption{Comparison in terms of training energy consumption ($x$-axis) and accuracy after the final incremental step ($y$-axis) across benchmarks and task sequence lengths.}
    \label{fig:energy_acc}
\end{figure}

\minisection{How do CL approaches compare in training energy consumption?} Figure~\ref{fig:energy_acc} gives a high-level overview of the relationship between training energy consumption (on the $x$-axis) and accuracy (on the $y$-axis) at the end of class-incremental training. JointTraining clearly achieves the highest accuracy, but also that it consumes significantly more energy than all other methods.

\begin{wrapfigure}{r}{0.45\textwidth}
    \centering
    \vspace{-15pt}
    \includegraphics[width=0.40\textwidth]{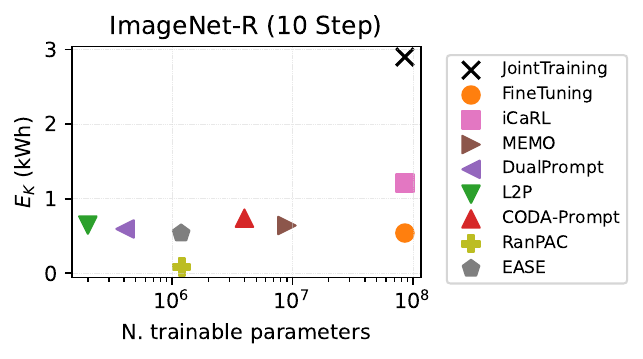}
    \caption{Trainable parameters versus Energy Consumption. SimpleCIL has zero parameters, while for RanPAC we select the number of trainable parameters for the first task.}
    \vspace{-10pt}
    \label{fig:trainable_params}
\end{wrapfigure}

Prompt-based methods like L2P, DualPrompt, and CODA-prompt achieve higher accuracy than FineTuning, while maintaining comparable energy costs despite having two orders of magnitude fewer trainable parameters, as shown in Figure~\ref{fig:trainable_params}. This challenges the common belief that the number of trainable parameters determines model efficiency, especially when starting from pre-trained models. The high energy consumption of prompt-based methods is due to the two forward passes for each backward pass: one for selecting prompts from the prompt pool, and another for making predictions based on the selected prompts.

Representation-based methods achieve the best balance between efficiency and accuracy (see Figure~\ref{fig:energy_acc}). As expected, SimpleCIL demonstrates very good energy efficiency and, thanks to the strong backbone, and a decent accuracy, outperforming FineTuning despite having no trainable parameters. RanPAC achieves optimal performance with minimal energy overhead by using a first task adaptation without further training on subsequent tasks.
These observations are consistent across all tested scenarios, indicating that the trade-off between consumption and accuracy does not change when varying benchmarks. %However, Figure~\ref{fig:energy_acc} provides only a partial perspective, as it does not explore how energy consumption scales with an increasing number of tasks.
\begin{figure}[t]
    \centering
    \includegraphics[width=0.9\textwidth]{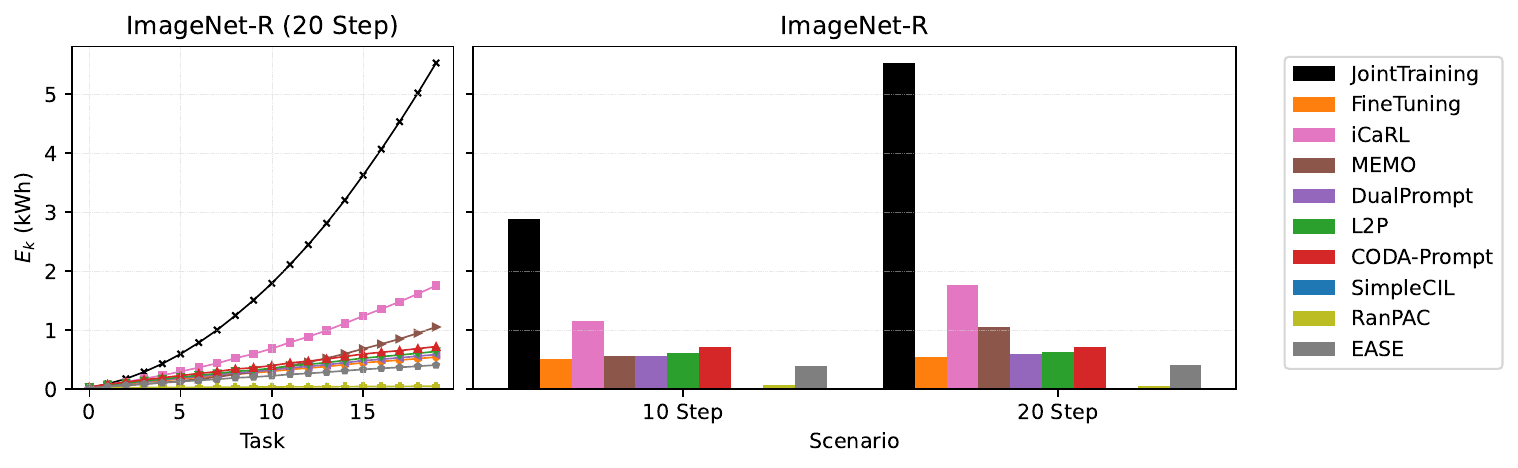}
    \caption{\textbf{Left}: Cumulative training energy consumption increases linearly for exemplar-free methods as the number of tasks grows, while exemplar-based methods show quadratic growth. \textbf{Right}: Total training energy consumption for each CL strategy on ImageNet-R split into 10 and 20 tasks, thus indicating fewer or more samples per task. Exemplar-based methods consume more energy with smaller tasks, whereas exemplar-free methods' energy consumption remains independent of task size.}
    \label{fig:cumulative_energy}
    \vspace{10pt}
    \includegraphics[width=0.9\textwidth]{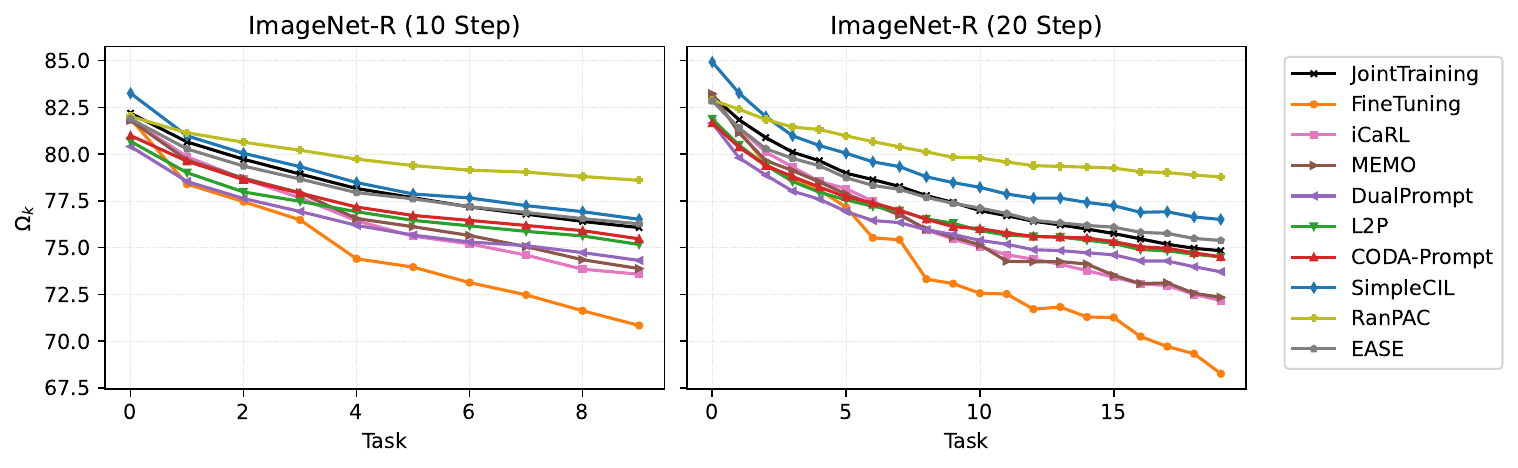}
    \caption{The Energy NetScore $\Omega_k$ across incremental steps on ImageNet-R. RanPAC consistently outperforms all others in both scenarios and on each task. Exemplar-based models perform well initially but become sub-optimal with increasing updates.}
    \label{fig:omega_score}
\end{figure}

\minisection{How do training energy costs scale with data and model updates?}
Figure~\ref{fig:energy_acc} provides only a partial perspective, as it does not explore how energy consumption scales with an increasing number of tasks. Figure~\ref{fig:cumulative_energy} completes the picture by showing on the left plot that all exemplar-free methods exhibit \textit{linear} growth in the cumulative energy consumption, in contrast to Joint Training, iCaRL and MEMO, which show \textit{quadratic} growth with the number of tasks, though in different manners. This observation is crucial, indicating that for a very large (potentially infinite) sequence of tasks, exemplar-based methods and in particular JointTraining may incur an unsustainable cost. Moreover, the bar plot on the right of Figure~\ref{fig:cumulative_energy} illustrates the effect of splitting the \textit{same} dataset into more tasks, i.e., halving the task sizes. The final energy consumption remains unchanged for all exemplar-free methods, but increases for the others.

%Among the methods that do not utilize memory from previous tasks, energy consumption increases linearly with different slopes. For a clearer comparison, Figure~\ref{fig:xxx} focuses only on exemplar-free methods. These methods cluster into two groups: those with higher energy demands effectively train parts of the network during incremental tasks, while SimpleCIL and RanPAC handle incremental training in a zero-shot manner. The only distinction between the latter two is that RanPAC performs an adaptation on the first task.

\minisection{How do CL approaches compare in Energy NetScore?}
The results of the experiments on CIFAR-100 and ImageNet-R are summarized in Table~\ref{tab:training_results}, which includes the Energy NetScore $\Omega_K$, defined in Equation~\ref{eqn:our_netscore} at the end of training. According to the Energy NetScore metric, the representation-based methods (i.e., SimpleCIL, EASE and RanPAC) significantly outperform the competitors. For relatively small benchmarks and short task sequences, exemplar-based methods, especially JointTraining, perform as well as or better than prompt-based methods. However, increasing the demand for model update on the same dataset, i.e., halving the task sizes in our experiments, the prompt-based methods close the gap with the exemplar-based methods. This trend is clearly shown in Figure~\ref{fig:omega_score}, where iCaRL initially scores competitively with the prompt-based methods, but as the number of exemplars during the training grows, its increased energy consumption lowers its score in the final tasks. A similar pattern can be observed when comparing EASE and JointTraining on ImageNet-R - 20 Step, where their positions in the ranking switch after 10 steps.
%\begin{figure}[t]
%    \centering
%    \includegraphics[width=\textwidth]{img/omega_inr.pdf}
%    \caption{The Energy NetScore $\Omega_k$ as a function of the number of steps on ImageNet-R. RanPAC consistently outperforms all other approaches in both scenarios and on each task along the sequences. Exemplar-based models perform well initially but become sub-optimal as model updates increase.}
%    \label{fig:omega_score}
%\end{figure}
\begin{table}[t]
\centering
\caption{Results on CIFAR100 and ImageNet-R. $A_K$, $E_K$ and $\Omega_K$ represent accuracy(\%), energy consumed(kWh), and Energy NetScore at the end of the task sequence, respectively. The memory column reports the number of exemplars employed.} 
\label{tab:training_results}
\setlength{\tabcolsep}{4pt}
\resizebox{0.85\textwidth}{!}
{%
\begin{tabular}{@{}lcccccccccc@{}}
\toprule
\multirow{2}{*}{\textbf{Method}} & \multirow{2}{*}{\textbf{Memory}} & \multicolumn{3}{c}{\textbf{CIFAR100 (10 Step)}} & \multicolumn{3}{c}{\textbf{ImageNet-R (10 Step)}} & \multicolumn{3}{c}{\textbf{ImageNet-R (20 Step)}} \\
             &  & $A_K (\uparrow)$    & $E_K (\downarrow)$     & $\Omega_K(\uparrow)$ & $A_K (\uparrow)$    & $E_K (\downarrow)$     & $\Omega_K(\uparrow)$ & $A_K (\uparrow)$    & $E_K (\downarrow)$     & $\Omega_K(\uparrow)$ \\ 
\cmidrule(r){1-1}
\cmidrule(lr){2-2}
\cmidrule(lr){3-5} 
\cmidrule(lr){6-8} 
\cmidrule(lr){9-11}

Joint Training                         & $\infty$ & 92.45 & 5.80 & 76.73    & 85.22 & 2.88 & 76.07    & 82.67 & 5.53 & 74.84   \\
Fine Tuning                            & \xmark & 69.09 & 1.07 & 73.50    & 56.68 & 0.52 & 70.85    & 48.98 & 0.55 & 68.25   \\
\cmidrule(r){1-1}
\cmidrule(lr){2-2}
\cmidrule(lr){3-5} 
\cmidrule(lr){6-8} 
\cmidrule(lr){9-11}
iCaRL~\cite{rebuffi2017icarl}          & 20/class & 80.55 & 1.61 & 75.73    & 69.72 & 1.15 & 73.58    & 66.00 & 1.76 & 72.17    \\
MEMO~\cite{zhou2022model}              & 20/class & 78.90 & 0.77 & 76.17    & 67.82 & 0.56 & 73.88    & 64.58 & 1.06 & 72.34    \\
L2P~\cite{wang2022learning}            & \xmark & 84.78 & 1.26 & 76.88    & 73.47 & 0.62 & 75.16    & 70.87 & 0.64 & 74.50    \\
DualPrompt~\cite{wang2022dualprompt}   & \xmark & 83.18 & 1.16 & 76.64    & 69.58 & 0.57 & 74.31    & 67.35 & 0.59 & 73.71    \\
CODA-Prompt~\cite{smith2023coda}       & \xmark & 85.73 & 1.48 & 76.90    & 75.42 & 0.71 & 75.47    & 71.45 & 0.72 & 74.52   \\
SimpleCIL\cite{zhou2023revisiting}     & \xmark & 76.12 & \textbf{0.02} & \underline{79.53}    & 61.35 & \textbf{0.01} & \underline{76.51}    & 61.35 & \textbf{0.01} & \underline{76.51}   \\
EASE~\cite{zhou2024expandable}         & \xmark & \underline{87.11} & 0.79 & 77.86    & \underline{76.08} & 0.39 & 76.27    & \underline{72.50} & 0.41 & 75.38   \\
RanPAC~\cite{mcdonnell2024ranpac}      & \xmark & \textbf{90.69} & \underline{0.12} & \textbf{80.60} & \textbf{78.07}  & \underline{0.07} & \textbf{78.59} & \textbf{77.48}  & \underline{0.05} & \textbf{78.82}   \\ 
\bottomrule
\end{tabular}
}
\end{table}

\begin{figure}
    \centering
    \begin{minipage}[b]{0.45\textwidth}
    \centering
    \resizebox{0.85\textwidth}{!}
    {%
        \begin{tabular}{lcccc}
            \toprule
            \multirow{2}{*}{\textbf{Method}} & \multirow{2}{*}{\textbf{Memory}} & \multicolumn{3}{c}{\textbf{DN4IL - 6 Step}} \\
                         &  & $A_K (\uparrow)$    & $E_K (\downarrow)$     & $\Omega_K(\uparrow)$  \\ 
            \cmidrule(r){1-1}
            \cmidrule(lr){2-2}
            \cmidrule(lr){3-5} 
            JointTraining                          & $\infty$ & 76.04 & 4.78 & 73.54  \\
            Fine Tuning                            & \xmark & 30.68 & 1.43 & 59.09  \\
            \cmidrule(r){1-1}
            \cmidrule(lr){2-2}
            \cmidrule(lr){3-5} 
            iCaRL~\cite{rebuffi2017icarl}           & 20/class & \underline{66.25} & 2.64 & 71.79  \\
            MEMO~\cite{zhou2022model}               & 20/class & 62.37 & 1.16 & 71.64  \\
            L2P~\cite{wang2022learning}             & \xmark & 40.37 & 1.68 & 63.62  \\
            DualPrompt~\cite{wang2022dualprompt}    & \xmark & 37.78 & 1.55 & 62.61  \\
            CODA-Prompt~\cite{smith2023coda}        & \xmark & 41.02 & 1.96 & 63.79  \\
            SimpleCIL~\cite{zhou2023revisiting}     & \xmark & 57.99 & \textbf{0.03} & \underline{74.34}  \\
            EASE~\cite{zhou2024expandable}          & \xmark & 50.74 & 1.05 & 68.16  \\
            RanPAC~\cite{mcdonnell2024ranpac}       & \xmark & \textbf{67.85} & \underline{0.20} & \textbf{75.01}  \\ 
            \bottomrule
        \end{tabular}
    }
        \captionof{table}{Results DN4IL. $A_K$, $E_K$ and $\Omega_K$ represent accuracy, energy consumed, and Energy NetScore at the end of the task sequence, respectively.}
        \label{tab:domain_res}
    \end{minipage}
    \hfill
    \begin{minipage}[b]{0.45\textwidth}
        \centering
        \includegraphics[width=0.9\textwidth]{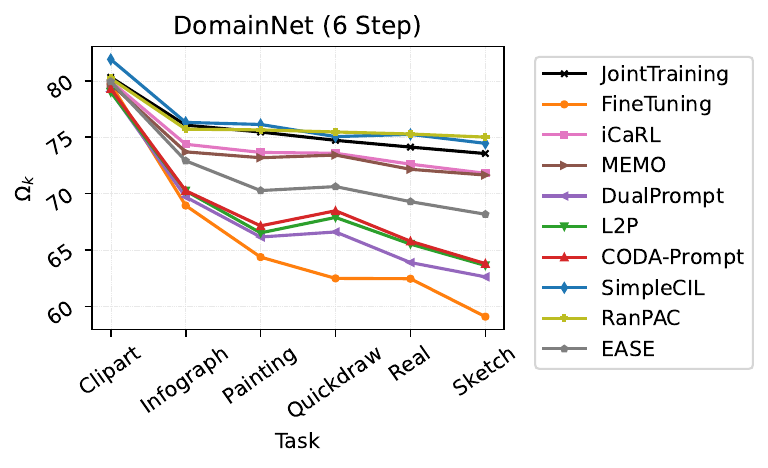} % replace 'example-image' with your image file
        \captionof{figure}{The Energy NetScore $\Omega_k$ as a function of the incremental step. Each step represent a different domain, as specified on the $x$-axis.}
        \label{fig:domain_omega}
    \end{minipage}
\end{figure}

\minisection{Domain and class-IL experiments.} 
We conducted an additional experiment in a challenging scenario, where each task involves changes in classes and in the domain. This setting better reflects real-world applications, where different objects may be observed under varying external conditions or using different instruments. Figure~\ref{fig:domain_omega} shows that representation-based methods, particularly SimpleCIL and RanPAC, continue to offer the best trade-off between accuracy and energy consumption. However, compared to previous scenarios, the accuracy gap with exemplar-based methods narrows, while it widens with JointTraining (see Table~\ref{tab:domain_res}).  For instance, comparing with the 10-step ImageNet-R experiment, the performance gap between RanPAC and iCaRL narrows from 9\% to 1.5\% points, while the gap with JointTraining widens from 5\% to 10\% points. This highlights the growing criticism of methods that either do not adapt the backbone or only use first-session adaptation, as they fail to fully address the challenges of CL~\cite{thede2024reflecting}. Such methods may be inadequate in scenarios where the context differs significantly from pre-training or where the domain gap between tasks shifts abruptly. Finally, Prompt-based methods and EASE seem to suffer in this setting.

%\subsection{Inference consumption}
\minisection{How does the CL strategy impact inference consumption?}
\begin{figure}[t]
    \centering
    \includegraphics[width=0.9\textwidth]{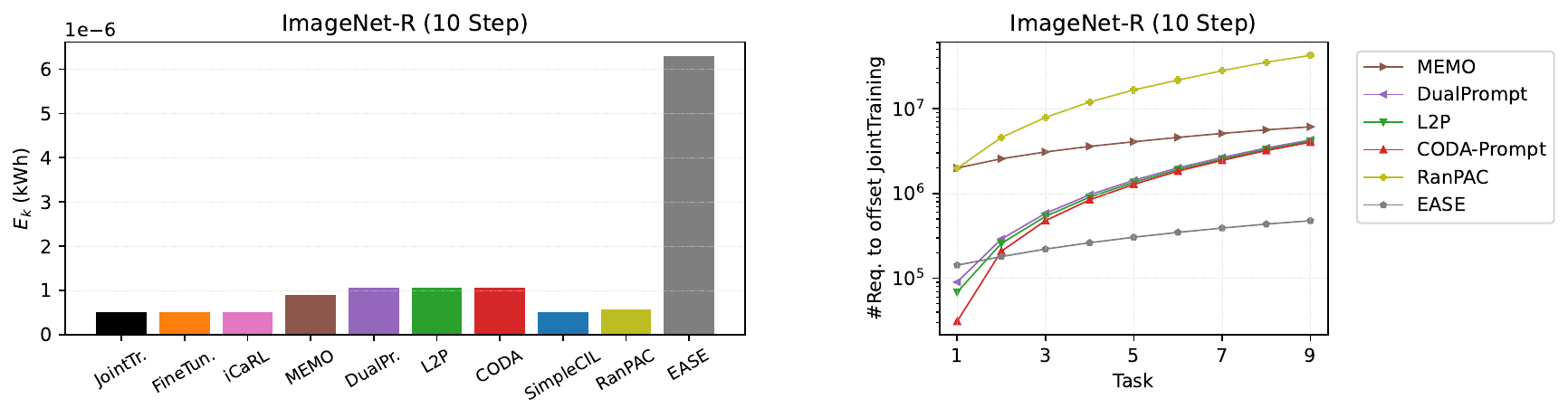}
    \caption{\textbf{Left}: Energy consumption for a single inference for each approach. \textbf{Right}: The number of requests required for approaches consuming more energy per inference, compared to JointTraining, to offset the energy savings gained during training.}
    \label{fig:inference_res}
\end{figure}
Once trained, machine learning models are used for inference, which can consume substantial energy due to high request volumes~\cite{JMLR:v24:23-0069}. Thus, assessing a model's environmental impact must consider efficiency during both training and inference. Figure~\ref{fig:inference_res} summarizes our findings regarding the inference energy consumption on ImageNet-R, with similar conclusions applicable to other benchmarks.

The bar plot on the left shows the energy consumption for a single inference (i.e., batch size of $1$) for each method. EASE has high inference consumption despite low training costs (see Table~\ref{tab:training_results}) as it repeats the inference along the backbone for each trained adapter, leading to linear scaling of the inference energy consumption with the number of tasks. Other methods, except MEMO that is a dynamic expandable architecture method, maintain constant energy consumption regardless of task sequence length. However, we observe that Prompt-based methods, despite minimal parameter overhead, consume roughly double the inference energy of other methods due to requiring a two forward pass for each prediction. In contrast, RanPAC, which only requires a single forward pass, effectively controls energy consumption, adding only a small overhead from the random projection in a larger feature space. SimpleCIL, FineTuning, iCaRL, and JointTraining incur in no additional inference costs beyond a standard forward pass. Specifically, the energy impact of adding exemplars is limited to the training phase, and regularizers like Knowledge Distillation (KD), despite needing two forward passes during training, do not increase inference costs.

Observing that a subset of CL approaches, aimed at reducing training costs, consume more energy during inference compared to JointTraining, raises a natural question: After how many requests do CL methods with inference overhead consume as much energy as retraining the model each time on the entire dataset? In Figure~\ref{fig:inference_res} (right), we show the results of this analysis. Specifically, the $y$-axis represents the number of model requests after which the accumulated training savings up to task $k$ are offset by the inference overhead. We observe that RanPAC, which adds minimal consumption per inference, proves to be the most efficient among all approaches considered, reaching the break-even point with JointTraining after more than $10^7$ requests. Prompt-based methods reach the break-even point with JointTraining at roughly the same time, as they have similar inference consumptions. EASE, due to its linear increase in inference energy consumption, reaches the break-even point with JointTraining after significantly fewer requests compared to all other methods.

Note that these numbers depend significantly on the size of the training dataset. Higher energy usage during training increases the number of requests required for methods with additional inference cost to reach the break-even point with JointTraining. Nevertheless, we believe that a truly effective CL algorithm should not have additional overhead during inference compared to JointTraining.

\minisection{Comprehensive evaluation.}
%\begin{wrapfigure}{r}{0.45\textwidth}
%    \centering
%    \includegraphics[width=0.35\textwidth]{img/train_inference_plot_imagenetr.pdf}
%    \caption{Energy NetScore $\Omega_k$ as a function of the steps on ImageNet-R. Each step considers energy consumption from both training and 10,000 inferences when computing $\Omega_k$.}
%    \label{fig:train_inference}
%\end{wrapfigure}
Here, we test a realistic scenario where training and inference alternate during the incremental step. For this analysis,  we considered the energy consumed for both training and answering (after each training step) to 10,000 requests, a number comparable to the size of the training dataset.
\begin{wrapfigure}{r}{0.45\textwidth}
    \vspace{-20pt}
    \centering
    \includegraphics[width=0.35\textwidth]{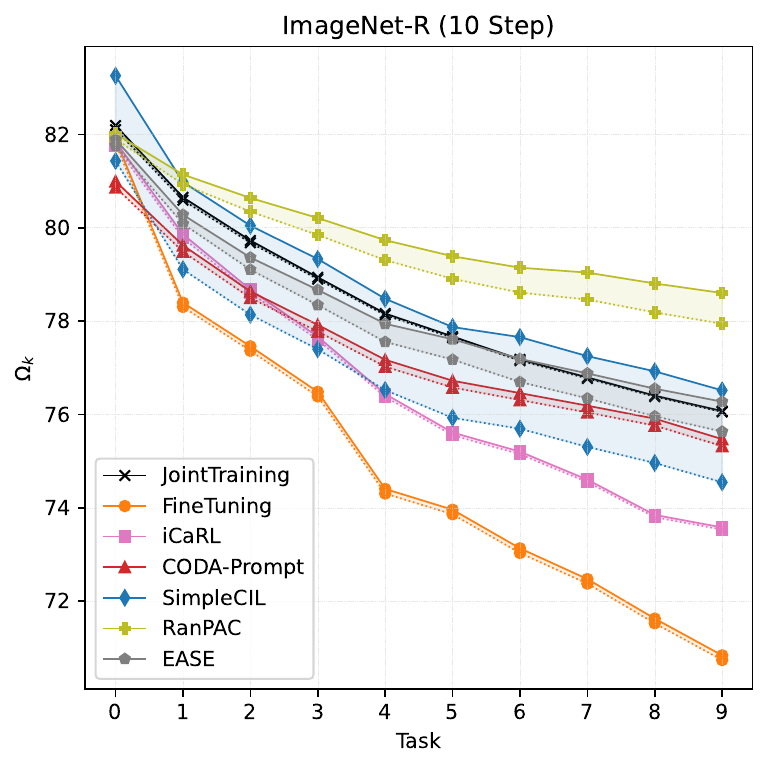}
     \vspace{-5pt}
    \caption{$\Omega_k$ as a function of the steps on ImageNet-R. Each step considers the consumption from both training and 10,000 inferences when computing $\Omega_k$.}
    \label{fig:train_inference}
     \vspace{-25pt}
\end{wrapfigure}
These consumption values were used to compute the Energy NetScore $\Omega_k$, shown in Figure~\ref{fig:train_inference}. The solid line represents the $\Omega_k$ score with zero inference cost, while the dashed line includes inference costs. SimpleCIL loses 2.5 points due to the impact of inference compared to its low training costs, while EASE loses about one point due to high inference costs. Despite a slight inference overhead, RanPAC remains the best solution in terms of $\Omega_k$. Other methods with higher training costs and relatively low inference costs, show minimal impact on their Energy NetScore from this number of requests.
%\begin{figure}[t]
%    \centering
%    \includegraphics[width=\textwidth]{img/train_inference_plot_all.pdf}
%    \caption{Energy NetScore $\Omega_k$ as a function of the steps on ImageNet-R. Each step considers energy consumption from both training and 10,000 inferences when computing $\Omega_k$.}
%    \label{fig:train_inference}
%\end{figure}

%%%%%%%%%%%%%%%%%%%%%%%%%%%%%%%%%%%%%%%%%%%%%%%%%%%%%%%%%%%%%%%%%%%% CONCLUSION

\section{Conclusion}
\emph{How green is continual learning, really?}  
In this paper we presented the first systematic analysis of the energy consumption of CL approaches with pre-trained backbones, and our results clearly indicate that \emph{it depends.} Our study highlights the complexities in selecting suitable CL algorithms, as these choices are highly context-dependent and influenced by factors such as the frequency of model updates, the magnitude of domain changes between steps, and the demands of inference. Nonetheless, our findings allow us to draw several general conclusions:
%\vspace{-15pt}
\begin{itemize}
    \item[\textbullet] Representation-based approaches, especially RanPAC, are the most energy-efficient during training while still maintaining performance close to JointTraining, as shown by the $\Omega_k$ values in Table~\ref{tab:training_results}. However, they still suffer when tested in large-domain shift scenarios (see Table~\ref{tab:domain_res}).
    \item[\textbullet] Exemplar-based methods are effective when the domain gap between steps is significant, performing comparably to representation-based approaches in $\Omega_k$ despite their quadratic growth in training energy consumption with an increasing number of tasks (see Table~\ref{tab:domain_res}).
    \item[\textbullet] The minimal overhead of RanPAC for inference makes it particularly appealing for real-word applications (see Figures~\ref{fig:inference_res} and~\ref{fig:train_inference}). In contrast, methods like EASE and MEMO, whose parameters and inference costs increase with the number of tasks, or prompt-based models, which require two forward passes per prediction, are less suitable for handling high volumes of requests. 
    \item[\textbullet] CL algorithms, to be competitive with JointTraining regardless of their potential usage, should not incur any additional computational overhead during inference.
\end{itemize}

\section*{Acknowledgements}
This work was supported by funding by the European Commission Horizon 2020 grant \#951911 (AI4Media).

\bibliographystyle{splncs04}
%\bibliography{main}

\newpage

\appendix
\section*{\huge Appendix}
\section{Hyperparameters and Architectural details}
\label{app:details}
For all approaches we used a ViT-B/16~\cite{dosovitskiy2021an} backbone with approximately 80 million parameters pretrained on ImageNet-21K and fine-tuned on ImageNet-1K. We followed the optimization settings and hyperparameters as suggested in PILOT~\cite{sun2023pilot} or in the original papers. To ensure a fair comparison between approaches, we maintained a fixed batch size of 64 and set the number of epochs to 20 for each task. \textbf{FineTuning}, \textbf{JointTraining}, and \textbf{iCaRL}~\cite{rebuffi2017icarl} do not introduce additional components to the architecture and all model parameters remain trainable, while iCaRL saves 20 exemplars per class. Similarly, \textbf{SimpleCIL}~\cite{zhou2023revisiting} does not add extra parameters; however, it freezes the entire backbone and adapts only the linear classifier during the incremental step. Below, we provide the specific architectural details for the other methods analyzed.

\minisection{JointTraining.} Since JointTraining is not included in the PILOT~\cite{sun2023pilot} framework, we implemented it from scratch. As previously mentioned, it does not introduce any additional trainable parameters beyond the default number for the ViT-B/16 model (approximately 80 million). We used SGD as the optimizer, with a momentum of 0.9, a weight decay of 0.0005, and a learning rate of 0.0001 for the backbone and 0.01 for the classifier. Like the other approaches, each task was trained for 20 epochs with a batch size of 64. The results we obtained for CIFAR-100 and ImageNet-R are consistent with those reported in~\cite{mcdonnell2024ranpac}.

\minisection{MEMO.} At each incremental step, MEMO~\cite{zhou2022model} keeps a task-agnostic feature extractor frozen and adds a task-specific module, introducing approximately 7 million new trainable parameters per step (see Figure~\ref{fig:trainable_params}). As a result, the model's size and inference computation cost increase with each additional task (see Figure~\ref{fig:inference_res}).

\minisection{Learning to Prompt.} As specified in Section~\ref{sec:cl_appr}, Learning to Prompt~\cite{wang2022learning} introduces learnable prompts that are prepended to the embeddings of the input data and trained during the incremental step, while keeping the backbone parameters frozen. Following the suggestions in~\cite{wang2022learning}, we use a prompt pool of size 10, with each prompt having a length of 5. For each query, the top 5 most aligned prompts are selected. With these choices, L2P adds less than 100,000 trainable parameters (see Figure~\ref{fig:trainable_params}), regardless of the number of steps.

\minisection{DualPrompt.} Similar to L2P, DualPrompt~\cite{wang2022dualprompt} introduces two different types of prompts that are prepended to the deep feature representation of the input data at different positions along the backbone and trained during the incremental step, while keeping the backbone parameters frozen. The general prompts are task-agnostic and are attached to the first two blocks of the backbone, while the expert prompts, which are task-specific, are attached to the next three blocks. In our experiment, the prompt length is fixed at 5 for both types. With these choices, DualPrompt adds approximately 300,000 trainable parameters (see Figure~\ref{fig:trainable_params}).

\minisection{CODA-Prompt.} As detailed in Section~\ref{sec:cl_appr}, CODA-Prompt~\cite{smith2023coda} modifies the concept of prompts by introducing the idea of \textit{prompt components}. Instead of selecting prompts from a predefined pool, it learns a set of prompt components that are combined through weighted summation. These weights are determined by a novel prompt-query matching process, enhanced by an attention mechanism. Following the original paper's recommendations, our experiment used a pool of 100 prompt components, each with a length of 8. The aggregated prompts are injected into the first five blocks of the backbone, similar to DUALPrompt. With these settings, CODA-Prompt adds approximately 3 million trainable parameters (see Figure~\ref{fig:trainable_params}) while keeping the backbone parameters frozen.

\minisection{EASE.} At each incremental step, EASE~\cite{zhou2024expandable} adds a task adapter while keeping the backbone frozen. Each adapter contains a bottleneck module for each of the 12 attention blocks in the ViT-B/16. These bottleneck modules contain a down-projection layer, an activation function, and an up-projection layer. Following the PILOT implementation, the down-projection reduces the feature dimension from 768 to 64, and the up-projection restores it. This setup adds approximately 100,000 new trainable parameters per attention block, totaling over 1 million per task (see Figure~\ref{fig:trainable_params}). Similar to MEMO, the model's size and inference computation costs increase with each additional task (see Figure~\ref{fig:inference_res}).

\minisection{RanPAC.} As outlined in Section~\ref{sec:cl_appr}, RanPAC~\cite{mcdonnell2024ranpac} is a representation-based method that performs an initial task adaptation and applies a random projection to the feature representations before classification. During the first task, instead of fully fine-tuning the backbone, it employs Parameter-Efficient Transfer Learning (PETL), adding around 1 million new trainable parameters (see Figure~\ref{fig:trainable_params}). Indeed, similar to EASE, RanPAC adds a bottleneck module for each of the 12 attention blocks in the ViT-B/16 model during the first session, introducing about 100,000 new trainable parameters per block. For subsequent tasks, all model parameters are frozen, with only the Gram matrix used for classification being updated. While the random projection layer introduces an additional 10 million parameters, these are not trainable.

\end{document}